# Accurate Uncertainties for Deep Learning Using Calibrated Regression

**Volodymyr Kuleshov** [1 2] **Nathan Fenner** [2] **Stefano Ermon** [1]

## Abstract

Methods for reasoning under uncertainty are a key building block of accurate and reliable machine learning systems. Bayesian methods provide a general framework to quantify uncertainty. However, because of model misspecification and the use of approximate inference, Bayesian uncertainty estimates are often inaccurate — for example, a 90% credible interval may not contain the true outcome 90% of the time. Here, we propose a simple procedure for calibrating any regression algorithm; when applied to Bayesian and probabilistic models, it is guaranteed to produce calibrated uncertainty estimates given enough data. Our procedure is inspired by Platt scaling and extends previous work on classification. We evaluate this approach on Bayesian linear regression, feedforward, and recurrent neural networks, and find that it consistently outputs well-calibrated credible intervals while improving performance on time series forecasting and model-based reinforcement learning tasks.

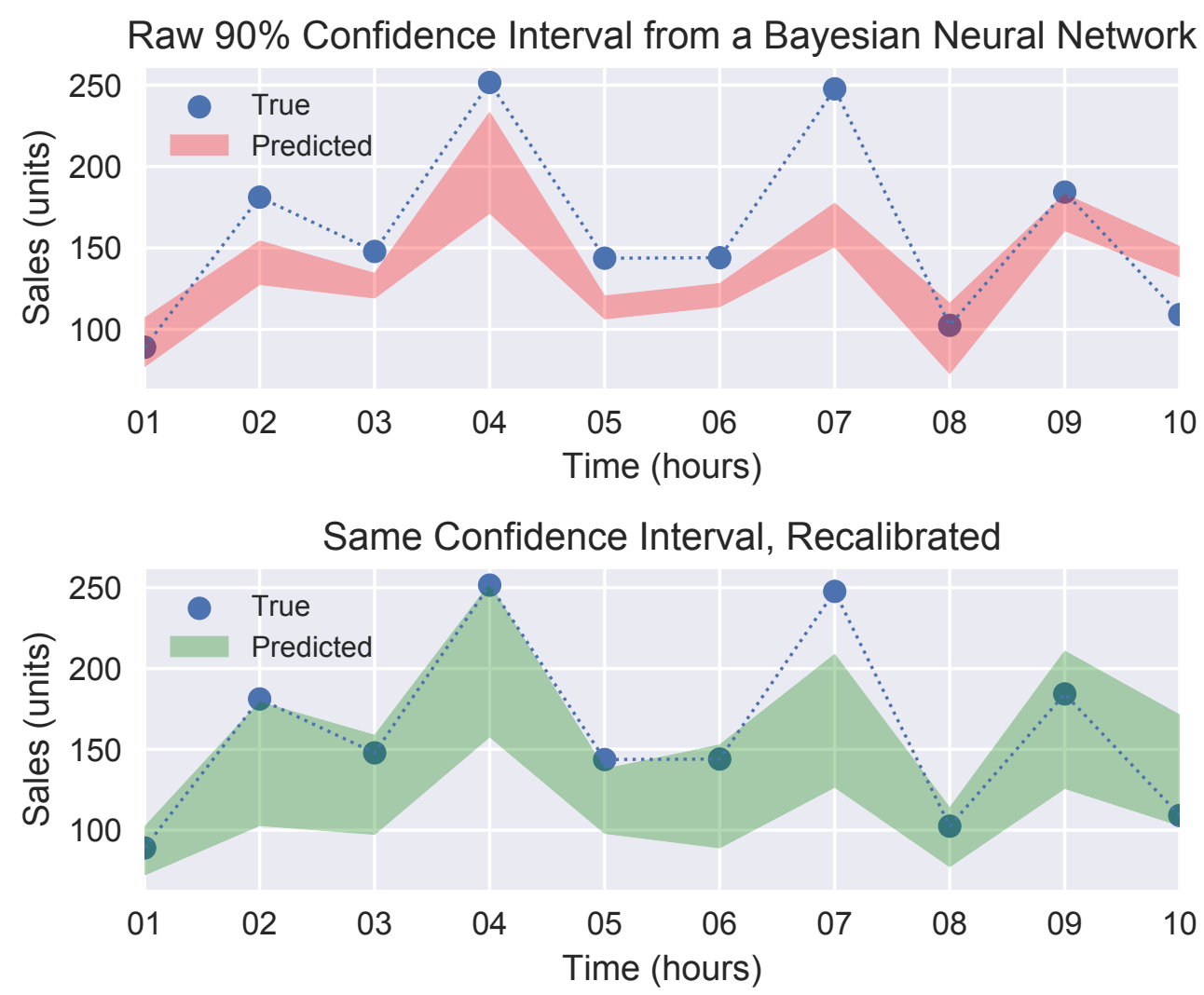

*Figure 1.* Top: Time series forecasting using a Bayesian neural network. Because the model is Bayesian, we may obtain a 90% credible interval around the forecast (red). However, the interval fails to capture the true data distribution: most points fall outside of it. Bottom: We propose a recalibration method that enables the original model to output a 90% credible interval (green) that correctly contains 9/10 points.

## 1. Introduction

Methods for reasoning and making decisions under uncertainty are an important building block of accurate, reliable, and interpretable machine learning systems. In many applications — ranging from supply chain planning to medical diagnosis to autonomous driving — faithfully assessing uncertainty can be as important as obtaining high accuracy. This paper explores uncertainty estimation over continuous variables in the context of modern deep learning models.

Bayesian approaches provide a general framework for dealing with uncertainty (Gal, 2016). Bayesian methods define a probability distribution over model parameters and derive uncertainty estimates by intergrating over all possible model weights. Recent advances in variational inference have greatly increased the scalability and usefulness of these approaches (Blundell et al., 2015).

In practice, however, Bayesian uncertainty estimates often fail to capture the true data distribution (Lakshminarayanan et al., 2017) — e.g., a 90% posterior credible interval generally does not contain the true outcome 90% of the time (Figure 1). In such cases, we say that the model is miscalibrated. This problem arises because of model bias: a predictor may not be sufficiently expressive to assign the right probability to every credible interval, just as it may not be able to always assign the right label to a datapoint.

Recently, Gal et al. (2017) and Lakshminarayanan et al. (2017) proposed uncertainty estimation techniques for deep neural networks, which include ensemble methods, heteroscedastic regression, and concrete dropout. These methods require modifying the model and may not always produce perfectly calibrated forecasts. Calibration has been extensively studied in the weather forecasting literature

[1]Stanford University, Stanford, California [2]Afresh Technologies, San Francisco, California. Correspondence to: Volodymyr Kuleshov <kuleshov@cs.stanford.edu>.

*Proceedings of the 35th International Conference on Machine Learning*, Stockholm, Sweden, PMLR 80, 2018.

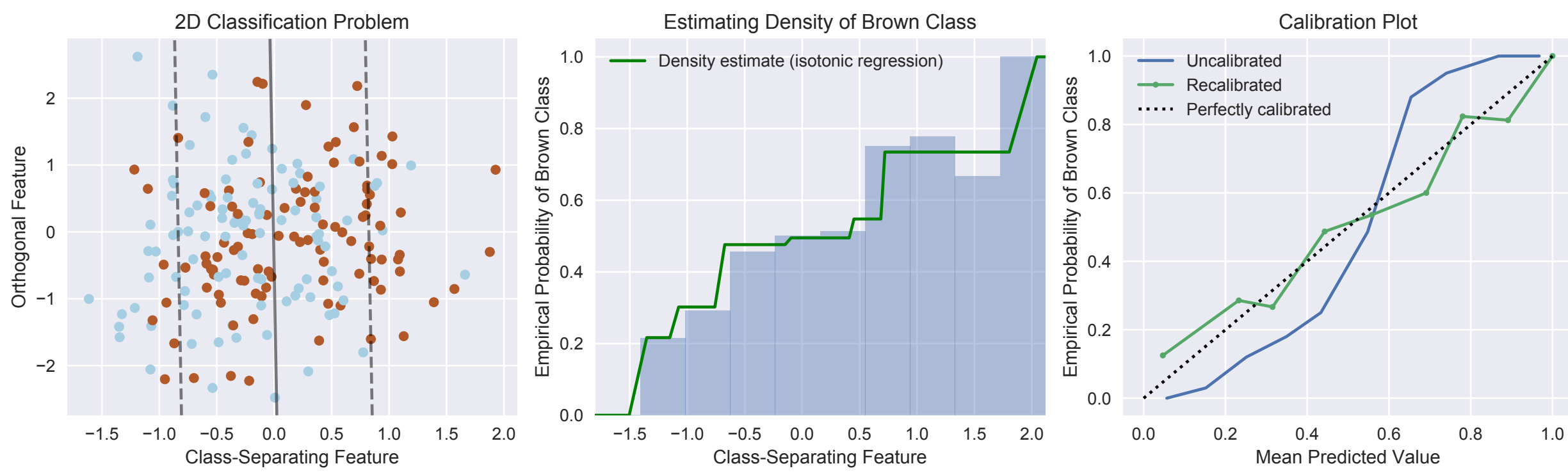

*Figure 2.* Calibrated classification. Left: Two classes are separated by a hyperplane in 2D. The $x$-axis is especially useful for separating the two classes. Middle: We project data onto the $x$-axis and fit a histogram (blue) or an isotonic regression model (green) to estimate the empirical probability of observing the brown class as a function of $x$. We may use these probabilities as approximately calibrated predictions. Right: The calibration of the original linear model and its recalibrated version are assessed by binning the predictions into ten intervals ($[0, 0.1], (0.1, 0.2], \dots$), and plotting the predicted vs. the observed frequency of the brown class in each interval.

(Gneiting and Raftery, 2005); however these techniques tend to be specialized and difficult to generalize beyond applications in climate science.

An alternative way to calibrate models has been explored in the support vector classification literature. These techniques — of which Platt scaling (Platt, 1999) is the most well-known — recalibrate the predictions of a pre-trained classifier in a post-processing step. As a result, these methods are classifier-agnostic and also typically very simple.

Here, we propose a new procedure for recalibrating any regression algorithm that is inspired by Platt scaling for classification. When applied to Bayesian and probabilistic deep learning models, it always produces calibrated credible intervals given a sufficient amount of i.i.d. data.

We evaluate our proposed algorithm on a range of Bayesian models, including Bayesian linear regression as well as feedforward and recurrent Bayesian neural networks. Our method consistently produces well-calibrated confidence estimates, which are in turn useful for several tasks in time series forecasting and model-based reinforcement learning.

**Contributions.** In summary, we introduce a simple technique for recalibrating the output of any regression algorithm, extending recalibration methods such as Platt scaling that were previously applicable only to classification. We then use this technique to solve an important problem in Bayesian deep learning: the miscalibration of credible intervals. We show that our results are useful in time series forecasting and in model-based reinforcement learning.

## 2. Calibrated Classification

This section is a concise overview of calibrated classification (Platt, 1999), and offers a reinterpretation of existing techniques that will be useful for deriving an extension to the regression and Bayesian settings in the next section.

**Notation.** We are given a labeled dataset $x_t, y_t \in \mathcal{X} \times \mathcal{Y}$ for $t = 1, 2, ..., T$ of i.i.d. realizations of random variables $X, Y \sim \mathbb{P}$, where $\mathbb{P}$ is the data distribution. Given $x_t$, a forecaster $H : \mathcal{X} \to (\mathcal{Y} \to [0, 1])$ outputs a probability distribution $F_t(y)$ targeting the label $y_t$. When $Y$ is continuous, $F_t$ is a cumulative probability distribution (CDF). In this section, we assume for simplicity that $\mathcal{Y} = \{0, 1\}$.

### 2.1. Calibration

Intuitively, calibration means that whenever a forecaster assigns a probability of $0.8$ to an event, that event should occur about 80% of the time. In binary classification, we have $\mathcal{Y} = \{0, 1\}$, and we say that $H$ is calibrated if

$$\frac{\sum_{t=1}^{T} y_t \mathbb{I}\{H(x_t) = p\}}{\sum_{t=1}^{T} \mathbb{I}\{H(x_t) = p\}} \to p \text{ for all } p \in [0, 1] \qquad (1)$$

as $T \to \infty$. Here, for simplicity, we use $H(x_t)$ to denote the probability of the event $y_t = 1$. When the $x_t, y_t$ are i.i.d. realizations of random variables $X, Y \sim \mathbb{P}$, a sufficient condition for calibration is:

$$\mathbb{P}(Y = 1 \mid H(X) = p) = p \text{ for all } p \in [0, 1]. \qquad (2)$$

**Calibration vs. Sharpness.** By itself, calibration is not enough to guarantee a useful forecast. For example, a forecaster that always predicts $\mathbb{E}[Y]$ is calibrated , but not very useful. Good predictions also need to be sharp, which intuitively means that probabilities should be close to zero or one. Note that an ideal forecaster is both calibrated and predicts outcomes with 100% confidence.

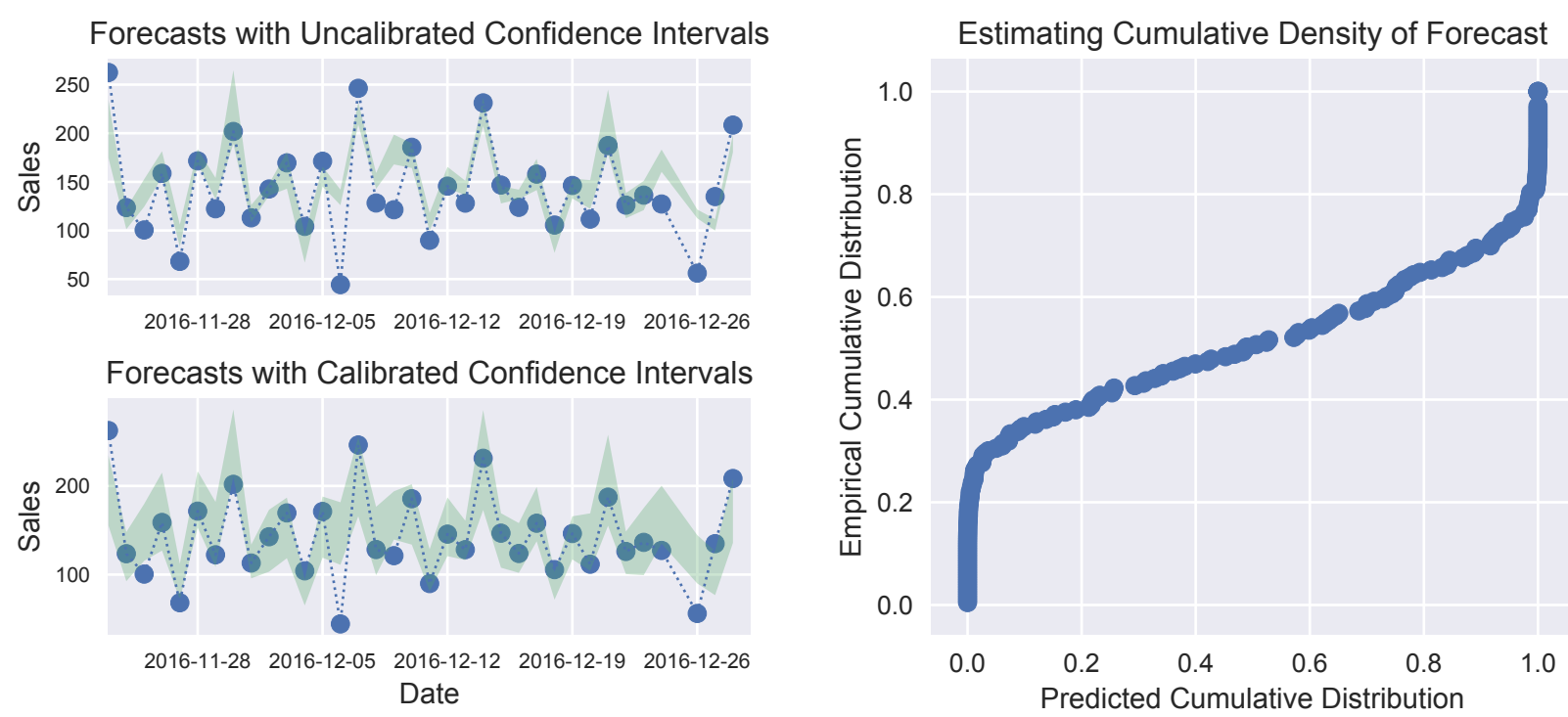

*Figure 3.* Calibrated regression. Left: A Bayesian neural network outputs probabilistic forecasts $F_t$ of future time series values $y_t$. The credible intervals do not always represent the true frequency of the prediction falling in the interval. Middle: For each credible interval, we plot the observed number of times the prediction falls in the interval (i.e. we estimate $\mathbb{P}(F_X(Y) \leq p)$). We fit this function and use it to output the actual probability of any given interval. Right: Forecast calibration can be assessed by plotting expected vs. empirical rates of observing an outcome $y_t$ in a set of ten intervals $(-\infty, F(p)]$ for $p = 0, 0.1, ..., 1$.

### 2.2. Training Calibrated Classifiers

Most classification algorithms — including logistic regression, Naive Bayes, and support vector machines (SVMs) — are not calibrated out-of-the-box. Recalibration methods train an auxiliary model $R : [0, 1] \to [0, 1]$ on top of a pre-trained forecaster $H$ such that $R \circ H$ is calibrated.

**Estimating a Probability Distribution.** When the $x_t, y_t$ are sampled i.i.d. from $\mathbb{P}$, choosing $R(p) = \mathbb{P}(Y = 1 \mid H(X) = p)$ yields a well-calibrated classifier $R \circ H$, according to the definition in Equation 2. Thus, recalibration can be framed as estimating the above conditional density.

Platt scaling (Platt, 1999) — one of the most widely used recalibration techniques — can be seen as approximating $\mathbb{P}(Y = 1 \mid H(X) = p)$ with a sigmoid (a valid assumption, in practice, when dealing with SVMs). Other recalibration methods fit this density with isotonic regression or kernel density estimation.

**Projections and Features.** A base classifier $H : \mathcal{X} \to \Phi$ may also output features $\phi \in \Phi \subseteq \mathbb{R}^d$ that do not correspond to probabilities. For example, an SVM outputs the margin between $x_t$ and the separating hyperplane. Such non-probabilistic $H$ can be similarly recalibrated by fitting $R : \Phi \to [0, 1]$ to $\mathbb{P}(Y = 1 \mid H(X) = \phi)$ (see Figure 2).

To gain further intuition, note that $H$ can be seen as projecting the $x_t$ into a low-dimensional space $\Phi$ (e.g., the SVM margin) such that the data is well separated in $\Phi$. The recalibrator $R : \Phi \to [0, 1]$ then performs density estimation to learn the Bayes-optimal classifier $\mathbb{P}(Y = 1 \mid H(X) = \phi)$. When $\phi$ is low-dimensional, this is tractable; furthermore $R \circ H$ is accurate because the classes $\mathcal{Y}$ are well-separated in $\phi$. Because $\mathbb{P}(Y = 1 \mid H(X) = \phi)$ is Bayes-optimal, $R \circ H$ is also calibrated.

**Diagnostic Tools.** The calibration of a classifier is typically assessed using calibration curves (Figure 2). Given a dataset $\{(x_t, y_t)\}_{t=1}^T$, let $p_t = H(x_t) \in [0, 1]$ be the forecasted probability. We group the $p_t$ into intervals $I_j$ for $j = 1, 2, ..., m$ that form a partition of $[0, 1]$ (e.g., $[0, 0.1]$, $(0.1, 0.2]$, etc.). A calibration curve plots the predicted average $p_j = T_j^{-1} \sum_{t:p_t \in I_j} p_t$ in each interval $I_j$ against the observed empirical average $p_j = T_j^{-1} \sum_{t:p_t \in I_j} y_t$, where $T_j = |\{t : p_t \in I_j\}|$. Perfect calibration corresponds to a straight line.

We can also assess sharpness by looking at the distribution of model predictions. When forecasts are sharp, most predictions are close to 0 or 1; unsharp forecasters make predictions closer to 0.5.

## 3. Calibrated Regression

In this section, we extend recalibration methods for classification to to regression ($\mathcal{Y} = \mathbb{R}$), and apply the resulting algorithm to Bayesian deep learning models. Recall that in regression, the forecaster $H$ outputs at each step $t$ a CDF $F_t$ targeting $y_t$. We will use $F_t^{-1} : [0, 1] \to \mathcal{Y}$ to denote the quantile function $F_t^{-1}(p) = \inf\{y : p \leq F_t(y)\}$.

### 3.1. Calibration

Intuitively, in a regression setting, calibration means than $y_t$ should fall in a 90% confidence interval approximately 90% of the time. Formally, we say that the forecaster $H$ is calibrated if

$$\frac{\sum_{t=1}^T \mathbb{I}\{y_t \leq F_t^{-1}(p)\}}{T} \to p \text{ for all } p \in [0, 1] \qquad (3)$$

as $T \to \infty$. In other words, the empirical and the predicted CDFs should match as the dataset size goes to infinity.

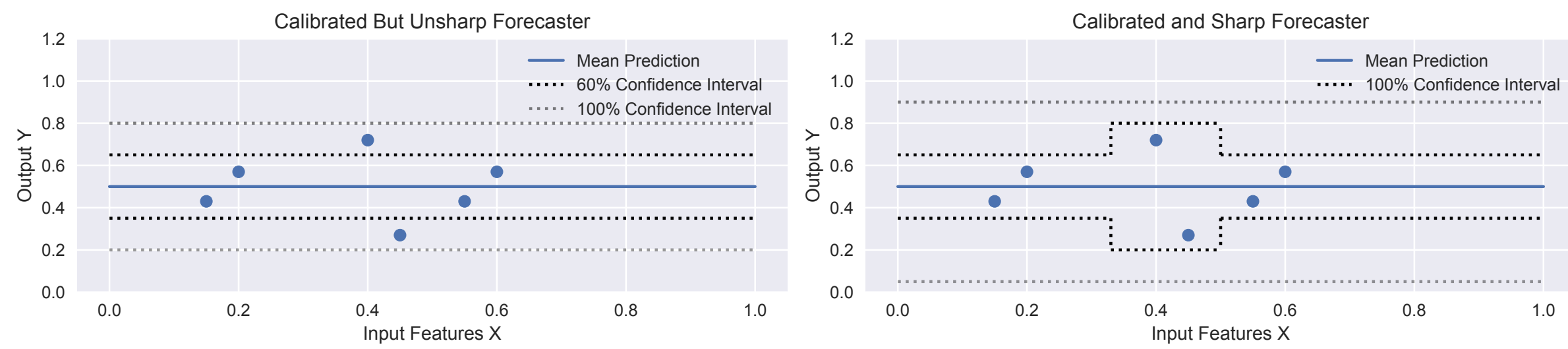

*Figure 4.* Calibrated forecasts with different sharpness. Left: A prediction (blue line) is surrounded by confidence intervals of uniform width; by counting the number of points falling in each interval, we may form calibrated confidence estimates. Right: A forecast with wider intervals in uncertain areas. The 100% confidence region is on average closer to the mean: this forecast is sharper and more useful.

When the $x_t, y_t$ are i.i.d. realizations of random variables $X, Y \sim \mathbb{P}$, a sufficient condition for this is

$$\mathbb{P}(Y \leq F_X^{-1}(p)) = p \text{ for all } p \in [0,1], \tag{4}$$

where we use $F_X = H(X)$ to denote the forecast at $X$. This formulation is related to the notion of probabilistic calibration of Gneiting et al. (2007).

Note that our definition also implies that

$$\frac{\sum_{t=1}^{T} \mathbb{I}\{F_t^{-1}(p_1) \leq y_t \leq F_t^{-1}(p_2)\}}{T} \to p_2 - p_1 \tag{5}$$

for all $p_1, p_2 \in [0,1]$ as $T \to \infty$. This extends our notion of calibration to general confidence intervals.

**Calibration and Sharpness.** As in classification, calibration by itself is not sufficient to produce a useful forecast. For example, it is easy to see that the forecast $F(y) = \mathbb{P}(Y \leq y)$ is calibrated; however it does even account for the features $X$ and thus cannot be accurate.

In order to be useful, forecasts must also be sharp. In a regression context, this means that the confidence intervals should all be as tight as possible around a single value. More formally, we want the variance $\text{var}(F_t)$ of the random variable whose CDF is $F_t$ to be small.

### 3.2. Training Calibrated Regression Models

We propose a simple recalibration scheme for producing calibrated forecasts that is closely inspired by classification techniques such as Platt scaling. Given a pre-trained forecaster $H$, we train an auxiliary model $R : [0,1] \to [0,1]$ such that the forecasts $R \circ F_t$ are calibrated (Algorithm 1).

This approach is simple, produces calibrated forecasts given enough i.i.d. data, and can be applied to any regression model, including recent Bayesian deep learning algorithms. Existing methods (Gal et al., 2017; Lakshminarayanan et al., 2017) require modifying the forecaster and may not produce calibrated forecasts even given large amounts of data.

---

**Algorithm 1** Recalibration of Regression Models.

**Input:** Uncalibrated model $H : \mathcal{X} \to (\mathcal{Y} \to [0,1])$ and calibration set $\mathcal{S} = \{(x_t, y_t)\}_{t=1}^T$.
**Output:** Auxiliary recalibration model $R : [0,1] \to [0,1]$.

1. Construct a recalibration dataset:

$$\mathcal{D} = \left\{ \Big( [H(x_t)](y_t), \hat{P}\left([H(x_t)](y_t)\right) \Big) \right\}_{t=1}^T,$$

where

$$\hat{P}(p) = |\{y_t \mid [H(x_t)](y_t) \leq p, t = 1, ..., T\}|/T.$$

2. Train a model $R$ (e.g., isotonic regression) on $\mathcal{D}$.

---

**Estimating a Probability Distribution.** Note that setting every forecast $F_t$ to $R \circ F_t$ where $R(p) := \mathbb{P}(Y \leq F_X^{-1}(p))$ yields a perfectly calibrated forecaster according to the definition in Equation 4. Thus, recalibration can be formulated as estimating the above cumulative probability distribution. This is similar to the classification setting, where we needed to estimate the conditional density $\mathbb{P}(Y = 1 \mid H(X) = p)$.

The intuition behind recalibration is that for any confidence level $p$, we may estimate from data the true probability $\mathbb{P}(Y \leq F_X^{-1}(p))$ of a random $Y$ falling in the credible region $(-\infty, F_X^{-1}(p)]$ below the $p$-th quantile of $F_X$. For example, we may count the fraction of points $(x_t, y_t)$ in a dataset that have this property or fit a regressor $R : [0,1] \to [0,1]$ to $\mathbb{P}(Y \leq F_X^{-1}(p))$, such that $R(p)$ estimates this probability for every $p$. Then, given a new forecast $F$, we may adjust the predicted probability $F(y)$ for the credible interval $(-\infty, y]$ to the true calibrated probability estimated empirically from data and given by $R \circ F(y)$. For example, if $p = 95\%$, but only 80/100 observed $y_t$ fall below the 95% quantile of $F_t$, then we adjust the 95% quantile to 80% (see Figure 3).

Specifically, given a dataset $\{(x_t, y_t)\}_{t=1}^T$, we may learn $\mathbb{P}(Y \leq F_X^{-1}(p))$ by fitting any regression algorithm to the recalibration set defined by $\{F_t(y_t), \hat{P}(F_t(y_t))\}_{t=1}^T$, where

$$\hat{P}(p) = \frac{|\{y_t \mid F_t(y_t) \leq p, t = 1, ..., T\}|}{T} \tag{6}$$

denotes the fraction of the data for which $y_t$ lies below the $p$-th quantile of $F_t$.

We recommend using isotonic regression (Niculescu-Mizil and Caruana, 2005) as the regression model on this dataset. This method accounts for the fact that the true function $\mathbb{P}(Y \leq F_X^{-1}(p))$ is monotonically increasing; it is also non-parametric, hence can learn the true distribution given enough i.i.d. data.

As in classifier recalibration, it is advisable to fit $R$ on a separate calibration set in order to reduce overfitting. Alternatively, one may break the data into $K$ folds, and train $K$ models in a way that is reminiscent of cross-validation: the hold-out fold serves as the calibration set and the model is trained on the remaining folds; at prediction time, the output is the average of the $K$ models.

### 3.3. Recalibrating Bayesian Models

Probabilistic forecasts $F_t$ are often obtained using Bayesian methods such as Bayesian neural networks or Gaussian processes. In practice, one often uses the mean and the variance $\mu, \sigma^2$ of dropout samples from a Bayesian neural network evaluated at $x_t$ to obtain a principled estimate of the predictive distribution over $y_t$ (Gal and Ghahramani, 2016a). The result is a probabilistic forecast $F_t(x_t)$ taking the form of a Gaussian $\mathcal{N}(\mu(x_t), \sigma^2(x_t))$.

However, if the true data distribution $\mathbb{P}(Y \mid X)$ is not Gaussian, uncertainty estimates derived from the Bayesian model will not be calibrated (see Figures 1 and 3). Algorithm 1 recalibrates uncertainty estimates from any black-box Bayesian model, making them accurate and useful.

### 3.4. Features for Recalibration

We may also use Algorithm 1 to recalibrate non-probabilistic forecasters, just as Platt scaling recalibrates SVMs. We may generalize the forecast to any increasing function $F(y) : \mathcal{Y} \to \Phi$ where $\Phi \subseteq \mathbb{R}$ defines a "feature" that correlates with the confidence of the classifier. We transform features into probability estimates by fitting a recalibrator $R : \Phi \to [0, 1]$ the following CDF:

$$\mathbb{P}(Y \leq F_X^{-1}(\phi)). \tag{7}$$

The simplest feature $\phi \in \Phi$ is the distance from the mean prediction, i.e. $[H(x)](y) = F_x(y) = y - \mu(x)$, where $\mu(x)$ is any point estimate of $Y$. Fitting $R$ essentially means counting the fraction of points that lie at any given distance of $\mu(x)$. Interestingly, this produces calibrated probabilistic forecasts even for an arbitrary (non-probabilistic) regressor $H$. However, confidence intervals will have the same width everywhere independently of $x$ (e.g. Figure 4, left); this makes them less useful at identifying points $x$ where the model is uncertain.

A better feature should account for uncertainty as a function of $x$. For example, we may use heteroscedastic regression to directly fit a mean and standard deviation $\mu(x), \sigma(x)$ and use $F_x(y) = (y - \mu(x))/\sigma(x)$. Combining features can further improve the sharpness of forecasts.

### 3.5. Diagnostic Tools

Next, we propose a set of diagnostic measures and visualizations in order to assess calibration and sharpness.

**Calibration.** We propose a calibration plot for regression inspired by the one for calibration. This plot displays the true frequency of points in each confidence interval relative to the predicted fraction of points in that interval.

More formally, we choose $m$ confidence levels $0 \leq p_1 < p_2 < \ldots < p_m \leq 1$; for each threshold $p_j$, we compute the empirical frequency

$$\hat{p}_j = \frac{|\{y_t \mid F_t(y_t) \leq p_j, t = 1, ..., T\}|}{T}. \tag{8}$$

To visualize calibration, we plot $\{(p_j, \hat{p}_j)\}_{j=1}^M$; calibrated forecasts correspond to a straight line. Note that for best results, the diagnostic dataset should be distinct from the calibration and training sets.

Finally, we propose using the calibration error as a numerical score describing the quality of forecast calibration:

$$\text{cal}(F_1, y_1, ..., F_T, y_T) = \sum_{j=1}^{m} w_j \cdot (p_j - \hat{p}_j)^2. \tag{9}$$

The scalars $w_j$ are weights. We used $w_j \equiv 1$ in our experiments; alternatively, choosing $w_j \propto |\{y_t \mid F_t(y_t) \leq p_j, t = 1, ..., T\}|$ decreases the importance of intervals that contain fewer data and that are more difficult to calibrate.

**Sharpness.** We propose measuring sharpness using the variance $\text{var}(F_t)$ of the random variable whose CDF is $F_t$. Low-variance predictions are tightly centered around one value. A sharpness score can be defined by

$$\text{sha}(F_1, ..., F_T) = \frac{1}{T} \sum_{t=1}^{T} \text{var}(F_t). \tag{10}$$

Note that this definition also applies to categorical variables; for a binary $Y$ with probability mass function $f$, we have $\text{var}(f) = f(1)(1 - f(1))$. The latter value is not only maximized at 0 or 1, but corresponds to the "refinement" term in the classical decomposition of the Brier score (Murphy, 1973).

| | Bayesian Linear Regression | | | Approx. Bayesian Neural Net | | | Concrete Dropout | | Deep Ensemble | |
|---|---|---|---|---|---|---|---|---|---|---|
| dataset | MAPE | Calibr. | Recal. | MAPE | Calibr. | Recal. | MAPE | Calibr. | MAPE | Calibr. |
| mpg | 0.107 | 0.053 | 0.057 | 0.091 | 0.102 | **0.021** | 0.081 | 0.068 | 0.079 | 0.087 |
| auto | 0.002 | 0.175 | 0.029 | 0.027 | 0.205 | **0.017** | 0.009 | 0.103 | 0.014 | 0.242 |
| crime | 0.086 | 0.030 | 0.016 | 0.086 | 0.070 | **0.015** | 0.084 | 0.182 | 0.082 | 0.050 |
| kinematics | 0.267 | 0.024 | **0.006** | 0.110 | 0.043 | 0.016 | 0.097 | 0.228 | 0.107 | 0.027 |
| stocks | 0.005 | 0.163 | 0.052 | 0.020 | 0.183 | **0.024** | 0.011 | 0.039 | 0.013 | 0.229 |
| cpu | 0.395 | 0.074 | **0.025** | 0.351 | 0.163 | 0.065 | 0.294 | 0.166 | 0.319 | 0.147 |
| bank | 0.572 | 0.073 | 0.083 | 0.395 | 0.134 | **0.057** | 0.410 | 0.177 | 0.390 | 0.126 |
| wine | 0.101 | 0.024 | **0.022** | 0.097 | 0.096 | 0.028 | 0.099 | 0.207 | 0.099 | 0.096 |

*Table 1.* Mean absolute percent error (MAPE) and calibration error (Equation 9) for two regression algorithms (Bayesian linear regression and a dense neural network) and two baselines. Recalibrating the regressors improves calibration and outperforms the baselines.)

## 4. Experiments

### 4.1. Setup

**Datasets.** We use eight UCI datasets varying in size from 194 to 8192 examples; examples carry between 6 and 159 continuous features. There is generally no standard train/test split, hence we randomly assign 25% of each dataset for testing, and use the rest for training. We report averages over 5 random splits. We also perform depth estimation on the larger Make3D dataset (Saxena et al., 2009), using the setup of Kendall and Gal (2017).

We also test our method on time series forecasting and reinforcement learning tasks. We use daily grocery sales from the Corporacion Favorita Kaggle dataset; we forecast the highest-selling item (#1503844) and use data from 2014-01-01 to 2016-05-31 in stores #1-4 for training and data from 2016-06-01 to 2016-12-31 for testing. We use autoregressive features from the past four days as well as binary indicators for the day of the week and the week of the year.

**Models.** The simplest model we study is Bayesian Ridge Regression (MacKay, 1992). The prior over the weights is a spherical Gaussian with a Gamma prior over the precision parameter. Posterior inference can be performed in closed form because the prior is conjugate.

We also consider feedforward and recurrent neural networks and we use the dropout approximation to variational inference of Gal and Ghahramani (2016a) to produce uncalibrated forecasts. In UCI experiments, the feedforward neural network has two layers of 128 hidden units with a dropout rate of 0.5 and parametric ReLU non-linearities. Recurrent networks are based on a standard GRU architecture with two stacked layers and a recurrent dropout of 0.5 (Gal and Ghahramani, 2016b). We use the DenseNet architecture of Jégou et al. (2017) for the depth regression task.

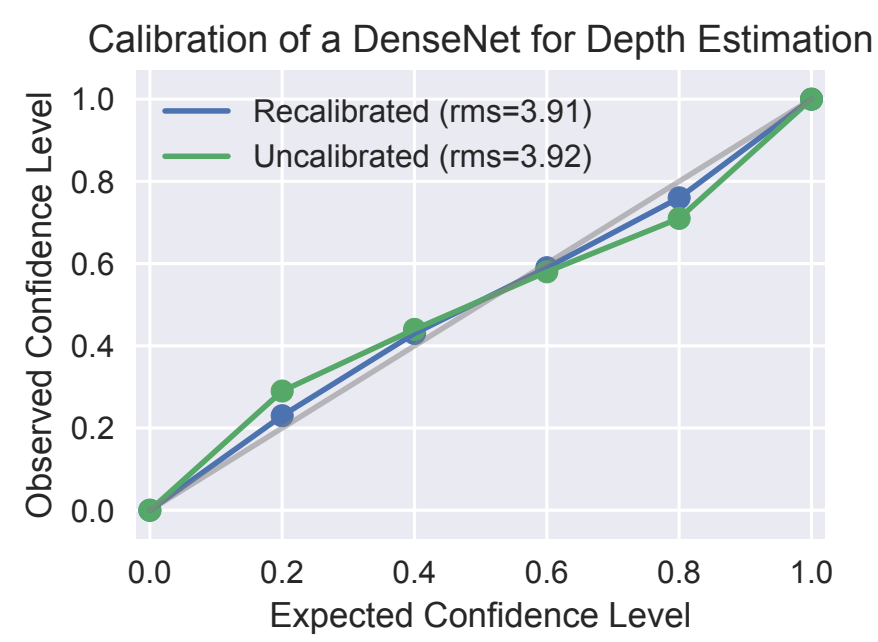

*Figure 5.* Calibration curves of a DenseNet model for depth estimation on the Make3D dataset, before and after calibration.

To perform recalibration, we first fit a base model on the training set, and then use isotonic regression as the recalibrator on the same training set. We didn't observe significant overfitting and did not use a distinct calibration set.

**Baselines.** We compare our approach against two recently proposed methods for improving the calibration of deep learning models: concrete dropout (Gal et al., 2017) and deep ensembles (Lakshminarayanan et al., 2017). Concrete dropout is a technique for learning the dropout probabilities based on the concrete distribution (Maddison et al., 2016); we use it to replace standard dropout in our neural network models. Discrete ensembles train multiple models with heteroscedastic regression and average their predictive distributions at runtime; in our experiments, we use an ensemble 5 instances of the same neural network that we use as the base predictor (except we add $\sigma(x)$ as an output).

### 4.2. UCI Experiments

Table 1 reports the accuracy (in terms of mean absolute percent error) and the test set calibration error (Equation 9) of Bayesian linear regression, a dense neural network, and two baselines on eight UCI datasets. We also report the re-calibrated error, which is significantly lower than that of the the original model. Concrete dropout and deep ensembles are often better calibrated than the regular neural network, but not as much as the recalibrated models. Recalibrated forecasts achieve similar accuracies to the baselines, even though the latter have more parameters.

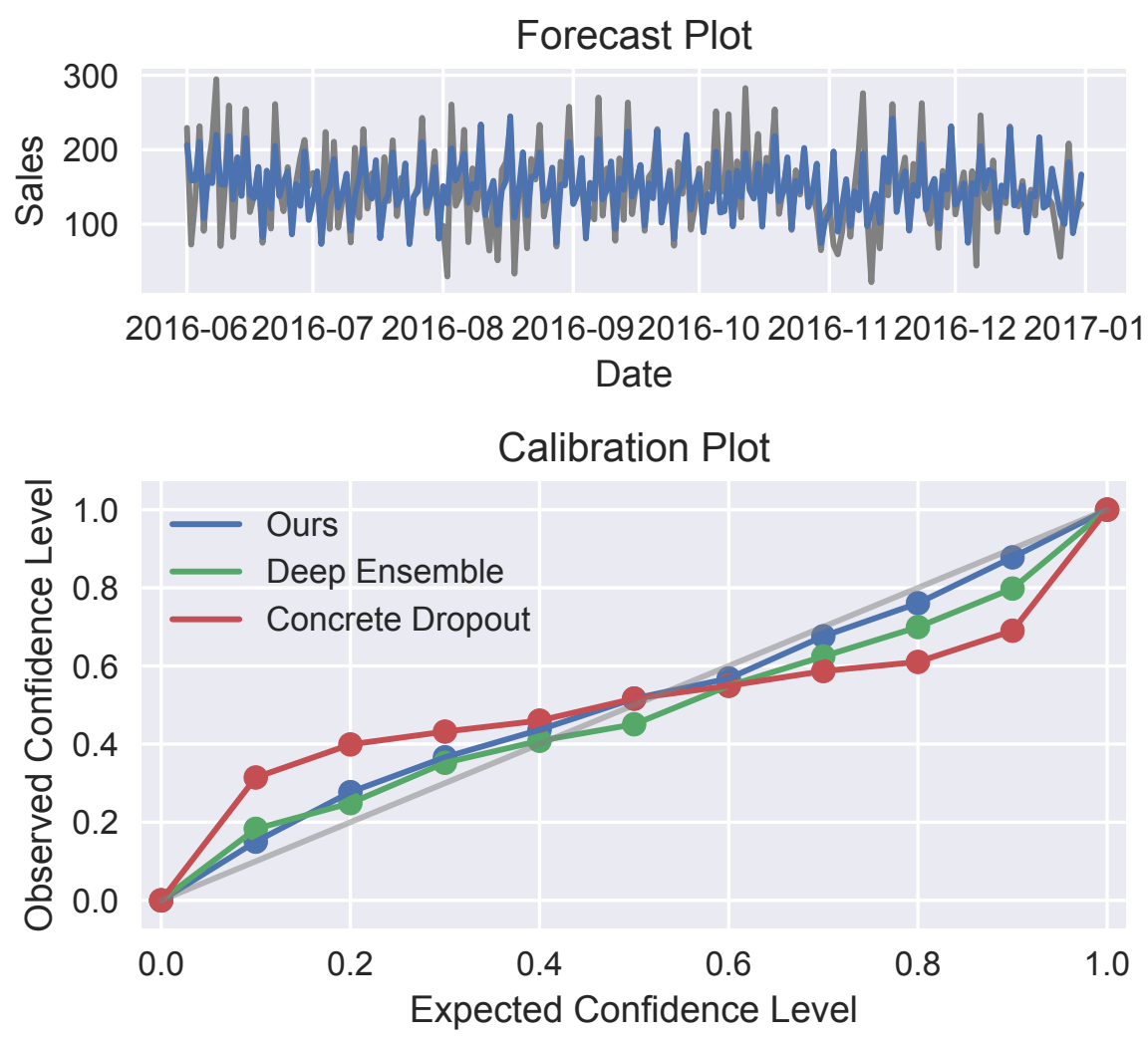

*Figure 6.* Time series forecasting. Top: Our recalibrated forecast (blue) predicts produce sales (gray). Bottom: Calibration plot comparing our method to two baselines.

### 4.3. Depth Estimation

We follow the setup of Kendall and Gal (2017). We compute per-pixel uncertainty estimates using dropout and measure calibration over all the individual pixels. We recalibrate on all the pixels in the training set. This yields an improvement in calibration error from 5.50E-02 to 2.41E-02, while preserving accuracy. The full calibration plot is given in Figure 5.

### 4.4. Time Series Forecasting

Next, we fit a recurrent network to forecast daily sales of item #1503844; we obtain mean absolute percent errors of 17.3-21.8% on the test set across the four stores. In Figure 3, we show that an uncalibrated 90% confidence interval misses most of the data points; however, the recalibrated confidence interval correctly contains about 90% of the true values.

Furthermore, we report in Figure 6 true and forecasted sales in store #1, as well as the calibration curves for both methods. The two baselines improve on the calibration of the original model, but our recalibration technique is the only one to achieve almost perfectly calibrated forecasts.

| Concrete Dropout | Deep Ensemble | Ours |
|---|---|---|
| \$60,082 | \$60,894 | \$61,690 |

*Table 2.* Cumulative reward (in dollars, averaged across 4 stores) of a model-based RL agent performing inventory management, using different algorithms to learn the state transition model.

### 4.5. Model-Based Reinforcement Learning

Uncertainty estimation is important in reinforcement learning to balance exploration and exploitation, as well as to improve model-based planning (Ghavamzadeh et al., 2015). Here, we focus on the latter task, and show how model-based reinforcement learning can be improved by calibrating the learned transition functions between states. Our task is inventory management, a classical application of reinforcement learning (Van Roy et al., 1997): on a set of days, an agent calculates order quantities for a perishable item in order to maximize store profits; transitions between states are defined by the probabilistic demand forecasts obtained in the previous section.

We formalize this task as a Markov decision process $(S, A, P, R)$. States $s \in S$ are sets of tuples $\{(q, \ell); \ell = 1, 2, ..., L\}$; each $(q, \ell)$ indicates that the store carries $q$ units of the item that expire in $\ell$ days ($L$ being the maximum shelf-life). Transition probabilities $P$ are defined through the following process: on each day the store sells $d$ units (a random quantity sampled from historical data not seen during training) which are removed from the inventory in $s$ (items leave in a first-in first-out manner); the shelf-life of the remaining items is decreased (spoiled items are thrown away). Actions $a \in A$ correspond to orders: the store receives $a$ items with a shelf life of $L$ before entering the next state $s'$. Finally, rewards are store profits, defined as sales revenue minus ordering costs.

We perform a simulation experiment on the grocery dataset: we use the Bayesian neural network from the previous section as our learned model of the environment (i.e., the state transition function). We then use dynamic programming with a 14-day horizon to determine the best action at each step. We evaluate the agent on the test portion of the Kaggle dataset, using the historical sales to define the state transitions. Item prices and costs are set to 1.99 and 1.29 respectively; items can be ordered three days a week in packs of 12 and arrive on the next day; the shelf-life of new items is always five days.

Table 2 shows the results. A calibrated state transition model allows the agent to better plan its actions and obtain a higher reward. This suggests that our method is useful for planning in model-based reinforcement learning.

## 5. Discussion

**Calibrated Bayesian Forecasts.** Our work proposes techniques for adjusting Bayesian models in a way that matches true empirical frequencies. This approach mixes frequentist and Bayesian ideas; a pure Bayesian would have instead defined a single model family and tried integrating over all possible models. Instead, we take an approach reminiscent of model criticism techniques (Box and Hunter, 1962) such as posterior predictive checking and its variants (Gelman and Hill, 2007; Kucukelbir et al., 2017). We fit a model to a dataset, and compare its predictions to real-world data, making adjustments if necessary.

Interestingly, our work suggests that a fully probabilistic model is not necessary to obtain confidence estimates: we may instead use simple features such as the distance from a point forecast. The advantage of more complex models is to provide a richer raw signal about their uncertainty, which may be recalibrated into sharper forecasts (Figure 4).

**Probabilistic Forecasting.** Gneiting et al. (2007) proposed several notions of calibration for continuous variables; our definition is most closely related to his concept of probabilistic calibration. The main difference is that Gneiting et al. (2007) define it relative to latent generative distributions, whereas our most general definition only considers empirical frequencies. We found that their notion of marginal calibration was too weak for our purposes, since it only preserves guarantees relative to the average distribution of the $y_t$ (which may be too high-variance to be useful).

Gneiting et al. (2007) also proposed that a probabilistic forecast should maximize sharpness subject to calibration. Interestingly, we take somewhat of an opposite approach: given a pre-trained model, we maximize for its calibration.

This form of recalibration preserves the accuracy of point estimates from the model. A recalibrated estimate of the median (or any other quantile) only becomes worse if there is insufficient data for recalibration, or there is a shift in the data distribution. However, the forecasts may become less sharp if the original forecaster was underestimating its uncertainty and producing credible intervals that were too tight around the mean.

**Applications.** Calibrated confidence estimates have been extensively studied by practitioners in medicine (Jiang et al., 2012), meteorology (Raftery et al., 2005), natural language processing (Nguyen and O'Connor, 2015), and other fields. Confidence estimates for continuous variables are important in computer vision applications, such as depth estimation (Kendall and Gal, 2017). Calibrated probabilities offer significant improvements over ordinary uncertainty estimates because they correspond to real-world empirical frequencies, and hence are interpretable.

## 6. Previous Work

**Calibrated Classification.** In the binary classification setting, Platt scaling (Platt, 1999) and isotonic regression (Niculescu-Mizil and Caruana, 2005) are effective and widely used to perform recalibration. They admit extensions to the multi-class setting (Zadrozny and Elkan, 2002) and to structured prediction (Kuleshov and Liang, 2015). They have also been studied in the context of modern neural networks (Guo et al., 2017; Gal et al., 2017; Lakshminarayanan et al., 2017). Recently Kuleshov and Ermon (2017) proposed methods that can quantify uncertainty via calibrated probabilities without any i.i.d. assumptions on the data, allowing inputs to potentially be chosen by a malicious adversary.

**Probabilistic Forecasting.** The study of calibration originates in the statistics literature (Murphy, 1973; Dawid, 1984), mainly in the context of evaluating probabilistic forecasts using proper loss functions (Gneiting and Raftery, 2007). Proper losses decompose into a calibration and sharpness term (Murphy, 1973); this decomposition also extends to probabilistic forecasts (Hersbach, 2000). Dawid (1984) also studied calibration in a Bayesian framework.

More recently, calibration has been studied in the literature on probabilistic forecasting, especially in the context of meteorology (Gneiting and Raftery, 2005). This resulted in specialized calibration systems (Raftery et al., 2005). Although most work on calibration focuses on classification, Gneiting et al. (2007) proposed several definitions of calibration for continuous variables. Their paper does not explore techniques for generating calibrated forecasts; we focus on the study of such algorithms in our work.

## 7. Conclusion

In summary, our paper formalized a notion of calibration for continuous variables, drawing close connections to work in calibrated classification. Inspired by these methods, we proposed a simple recalibration technique that generates calibrated probabilistic forecasts given enough i.i.d. data. Furthermore, we introduced visualizations and metrics for evaluating calibration and sharpness. Finally, we demonstrated the practical importance of calibration by applying our method to Bayesian neural networks. Our method consistently produces well-calibrated uncertainty estimates; this result is useful in time series forecasting, reinforcement learning, as well as more generally to construct reliable, interpretable, and interactive machine learning systems.

## References

Yarin Gal. *Uncertainty in Deep Learning*. PhD thesis, University of Cambridge, 2016.

Charles Blundell, Julien Cornebise, Koray Kavukcuoglu, and Daan Wierstra. Weight uncertainty in neural networks. *arXiv preprint arXiv:1505.05424*, 2015.

Balaji Lakshminarayanan, Alexander Pritzel, and Charles Blundell. Simple and scalable predictive uncertainty estimation using deep ensembles. *arXiv preprint arXiv:1612.01474*, 2017.

Yarin Gal, Jiri Hron, and Alex Kendall. Concrete dropout. In *Advances in Neural Information Processing Systems*, pages 3584–3593, 2017.

Tilmann Gneiting and Adrian E Raftery. Weather forecasting with ensemble methods. *Science*, 310(5746):248–249, 2005.

J. Platt. Probabilistic outputs for support vector machines and comparisons to regularized likelihood methods. *Advances in Large Margin Classifiers*, 10(3):61–74, 1999.

T. Gneiting, F. Balabdaoui, and A. E. Raftery. Probabilistic forecasts, calibration and sharpness. *Journal of the Royal Statistical Society: Series B (Statistical Methodology)*, 69(2):243–268, 2007.

A. Niculescu-Mizil and R. Caruana. Predicting good probabilities with supervised learning. In *Proceedings of the 22nd international conference on Machine learning*, pages 625–632, 2005.

Yarin Gal and Zoubin Ghahramani. Dropout as a Bayesian approximation: Representing model uncertainty in deep learning. In *Proceedings of the 33rd International Conference on Machine Learning (ICML-16)*, 2016a.

A. H. Murphy. A new vector partition of the probability score. *Journal of Applied Meteorology*, 12(4):595–600, 1973.

Ashutosh Saxena, Min Sun, and Andrew Y Ng. Make3d: Learning 3d scene structure from a single still image. *IEEE transactions on pattern analysis and machine intelligence*, 31(5): 824–840, 2009.

Alex Kendall and Yarin Gal. What uncertainties do we need in bayesian deep learning for computer vision? In *Advances in Neural Information Processing Systems*, pages 5580–5590, 2017.

David JC MacKay. Bayesian interpolation. *Neural computation*, 4(3):415–447, 1992.

Yarin Gal and Zoubin Ghahramani. A theoretically grounded application of dropout in recurrent neural networks. In *Advances in Neural Information Processing Systems 29 (NIPS)*, 2016b.

Simon Jégou, Michal Drozdzal, David Vazquez, Adriana Romero, and Yoshua Bengio. The one hundred layers tiramisu: Fully convolutional densenets for semantic segmentation. In *Computer Vision and Pattern Recognition Workshops (CVPRW), 2017 IEEE Conference on*, pages 1175–1183. IEEE, 2017.

Chris J Maddison, Andriy Mnih, and Yee Whye Teh. The concrete distribution: A continuous relaxation of discrete random variables. *arXiv preprint arXiv:1611.00712*, 2016.

Mohammad Ghavamzadeh, Shie Mannor, Joelle Pineau, Aviv Tamar, et al. Bayesian reinforcement learning: A survey. *Foundations and Trends® in Machine Learning*, 8(5-6):359–483, 2015.

Benjamin Van Roy, Dimitri P Bertsekas, Yuchun Lee, and John N Tsitsiklis. A neuro-dynamic programming approach to retailer inventory management. In *Decision and Control, 1997., Proceedings of the 36th IEEE Conference on*, volume 4, pages 4052–4057. IEEE, 1997.

George EP Box and William G Hunter. A useful method for model-building. *Technometrics*, 4(3):301–318, 1962.

Andrew Gelman and Jennifer Hill. *Data analysis using regression and multilevelhierarchical models*, volume 1. Cambridge University Press New York, NY, USA, 2007.

Alp Kucukelbir, Yixin Wang, and David M. Blei. Evaluating Bayesian models with posterior dispersion indices. In *Proceedings of the 34th International Conference on Machine Learning*, volume 70 of *Proceedings of Machine Learning Research*, pages 1925–1934, International Convention Centre, Sydney, Australia, 06–11 Aug 2017.

X. Jiang, M. Osl, J. Kim, and L. Ohno-Machado. Calibrating predictive model estimates to support personalized medicine. *Journal of the American Medical Informatics Association*, 19 (2):263–274, 2012.

Adrian E Raftery, Tilmann Gneiting, Fadoua Balabdaoui, and Michael Polakowski. Using bayesian model averaging to calibrate forecast ensembles. *Monthly weather review*, 133(5): 1155–1174, 2005.

K. Nguyen and B. O'Connor. Posterior calibration and exploratory analysis for natural language processing models. In *Empirical Methods in Natural Language Processing (EMNLP)*, pages 1587–1598, 2015.

B. Zadrozny and C. Elkan. Transforming classifier scores into accurate multiclass probability estimates. In *International Conference on Knowledge Discovery and Data Mining (KDD)*, pages 694–699, 2002.

V. Kuleshov and P. Liang. Calibrated structured prediction. In *Advances in Neural Information Processing Systems (NIPS)*, 2015.

Chuan Guo, Geoff Pleiss, Yu Sun, and Kilian Q Weinberger. On calibration of modern neural networks. *arXiv preprint arXiv:1706.04599*, 2017.

Volodymyr Kuleshov and Stefano Ermon. Estimating uncertainty online against an adversary. In *AAAI*, pages 2110–2116, 2017.

A. P. Dawid. Present position and potential developments: Some personal views: Statistical theory: The prequential approach. *Journal of the Royal Statistical Society. Series A (General)*, 147:278–292, 1984.

Tilmann Gneiting and Adrian E Raftery. Strictly proper scoring rules, prediction, and estimation. *Journal of the American Statistical Association*, 102(477):359–378, 2007.

Hans Hersbach. Decomposition of the continuous ranked probability score for ensemble prediction systems. *Weather and Forecasting*, 15(5):559–570, 2000.